\documentclass[final,5p,times,twocolumn]{elsarticle}

\usepackage{amssymb}
\usepackage{amsmath}

\usepackage[ruled,vlined,linesnumbered]{algorithm2e}
\usepackage[table]{xcolor}
\usepackage{threeparttable}
\usepackage{multirow}
\usepackage{booktabs}
\usepackage{makecell}
\biboptions{sort&compress}
\usepackage[pagebackref=false,breaklinks=true,letterpaper=true,colorlinks,bookmarks=false]{hyperref}

\journal{}

\begin{document}

\begin{frontmatter}

\title{Heterogeneous Element-Aware Cross-Version Differencing of Scientific Documents via Layout-Aware Alignment and Structure-Aware Reasoning}

\author[label1,label2]{Zhen Yin\corref{cor1}}
\author[label1]{Wenkang An}
\author[label1]{Hao Wang}
\author[label1]{Keran You}
\affiliation[label1]{organization={Beijing Renhe Information Technology Co., Ltd.},
            city={Beijing},
            postcode={100096}, 
            country={China}}
            
\affiliation[label2]{organization={Key Laboratory of Digital Publishing and Total Process Management of Scientific and Technical Journals},
            city={Beijing},
            postcode={100083}, 
            country={China}}

\cortext[cor1]{Corresponding author}

\begin{abstract}
Cross-version differencing of scientific documents is a fundamental task in scholarly publishing and technical document workflows, yet it remains challenging because scientific documents are page-structured artifacts composed of heterogeneous elements, including text, tables, formulas, figures, and layout cues. Existing text-sequence-based methods often lose layout and structural information, whereas image-based comparison methods lack semantic interpretability and are sensitive to rendering variation. To address these limitations, this paper proposes a layout-aware and heterogeneous element-aware framework for scientific document differencing. The proposed framework first decomposes each document version into semantically typed elements, then establishes cross-version correspondence through an alignment-first mechanism that jointly models spatial, content, and structural compatibility, and finally performs type-aware difference reasoning over aligned element pairs. Unlike conventional comparison pipelines, the framework supports unified change detection, change localization, structure-awareness analysis, and alignment/matching evaluation across text, tables, formulas, and figures. We evaluate the proposed method on real-world cross-version scientific PDF data collected from production proofreading workflows of journal editorial offices, covering diverse disciplines, layouts, proofreading stages, and revision patterns. Experimental results show that the proposed framework consistently outperforms element-specific baselines across heterogeneous document elements. It achieves detection F1 scores of 0.903, 0.855, 0.862, and 0.845 for text, tables, formulas, and figures, respectively, while also improving localization, structure-awareness, and matching quality. Ablation and sensitivity analyses further confirm the effectiveness of cross-version alignment, type-specific representations, structure-aware reasoning, and compatibility-weight design. These results demonstrate that heterogeneous element-aware differencing provides a more robust and interpretable solution for scientific document comparison in realistic editorial production scenarios.
\end{abstract}


\end{frontmatter}

\section{Introduction}
\label{introduction}

Cross-version differencing of scientific documents is intrinsically challenging because the compared objects are not homogeneous text sequences but heterogeneous page-structured artifacts composed of text, tables, formulas, figures, and layout cues \citep{pun2023survey,xia2024docgenome}. In scholarly publishing and technical documentation workflows, such comparison must go beyond identifying whether two document versions differ, and further determine where a revision occurs, which semantic element is affected, and whether the change is content-level, formatting-level, or structure-level. In practice, revisions may involve paragraph movement, table reformatting, formula relocation, figure replacement, caption adjustment, or typography-sensitive modifications such as bold/italic shifts, font-size variation, and superscript/subscript changes. These characteristics make reliable differencing difficult if the problem is reduced to either linear text comparison or whole-page visual matching alone \citep{hsu2022neural,chen2025xtragpt}.

Existing comparison paradigms remain insufficient for this setting. Sequence-based text diff methods are effective for homogeneous textual streams, but they tend to discard page structure, ignore heterogeneous element boundaries, and become unstable under reflow, reordering, or layout-sensitive revisions \citep{morikawa2024revtoken,santosh2024tale}. Image-based comparison methods can highlight page-level visual discrepancies, but they usually lack semantic interpretability and cannot reliably distinguish meaningful scientific revisions from layout noise \citep{park2025text,li2025document,feng2025dolphin}. At the same time, recent progress in document intelligence has mainly focused on element-specific tasks, such as layout analysis, table structure recognition, and mathematical expression modeling \citep{ouyang2025omnidocbench,sun2025pp}. Although these advances have substantially improved the understanding of individual document components, they do not directly address unified cross-version differencing for complex scientific documents.

This limitation is especially pronounced in scientific documents, where internal structure is often tightly coupled with semantic meaning. For tables, row–column organization and merged-cell relations are integral to interpretation, making structure-preserving representations essential \citep{zhong2019publaynet,yin2025enhancing}. For formulas, symbolic and tree-based representations are necessary because visually similar expressions may differ substantially in operator hierarchy and mathematical semantics \citep{wang2021scientific,ding2025tagging}. For text, typography-sensitive attributes such as font size, bold/italic style, and superscript/subscript status may also carry structural or semantic implications rather than merely reflecting visual formatting \citep{binmakhashen2019document,shi2025stepwise}. More broadly, document layout analysis has shown that spatially grounded representations are critical for identifying semantically meaningful regions on page-structured documents \citep{ma2026mamba}. These observations indicate that scientific document differencing should not be treated as a single homogeneous comparison problem, but rather as a layout-aware, heterogeneous, and structure-sensitive reasoning task \citep{wang2025unihdsa,qi2025yolo}.

\begin{figure}[t]
    \centering
    \includegraphics[width=1\linewidth]{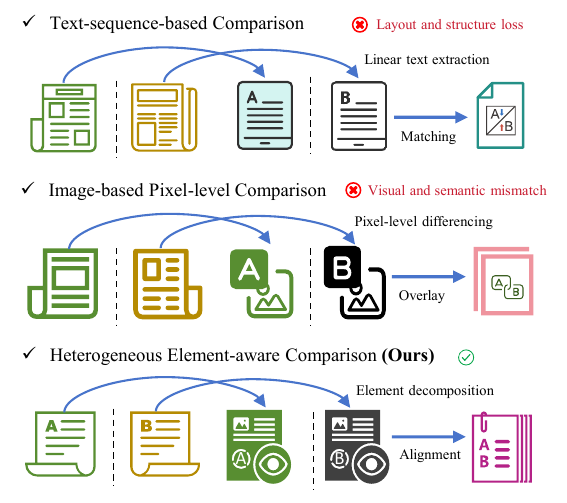}
\vspace{-.5cm}
\caption{
Comparison of mainstream document comparison paradigms and the proposed heterogeneous element-aware framework. Text-sequence-based methods often lose layout and structure information, whereas image-based methods are prone to visual ambiguity and limited semantic interpretability. The proposed framework instead compares heterogeneous structured elements for more robust and interpretable document-level differencing.}
\label{fig:Figure1}
\end{figure}

Motivated by this observation, we propose a layout-aware and element-specific framework for cross-version differencing of scientific documents. Instead of directly comparing text sequences or full-page images, the proposed method first decomposes each document into heterogeneous semantic elements, then establishes stable cross-version correspondence, and finally performs structure-aware difference reasoning on aligned element pairs. The framework follows three principles. First, it is layout-aware, because page structure is explicitly incorporated into element decomposition and alignment. Second, it is element-specific, because text, tables, formulas, and figures are represented and compared in type-appropriate spaces rather than under a single low-level representation. Third, it is alignment-first and structure-aware, because differencing is performed only after stable correspondence has been established and is further extended from change detection to localization and structural interpretation. As a result, the proposed framework reduces spurious differences caused by paragraph displacement, page reflow, and element relocation, while supporting fine-grained differencing over heterogeneous document elements and preserving typography-sensitive as well as structure-sensitive cues that are typically lost in generic text- or image-based pipelines. An overview of the proposed framework is illustrated in Fig.~\ref{fig:Figure1}.

The main contributions of this work are as follows:
\begin{itemize}
    \item We define cross-version scientific document differencing as a unified heterogeneous element-level task and establish an evaluation protocol covering detection, localization, structure awareness, and alignment quality.
    
    \item We propose a layout-aware cross-version alignment mechanism that jointly models spatial, content, and structural compatibility for stable heterogeneous element correspondence.
    
    \item We develop an element-specific structure-aware differencing framework for text, tables, formulas, and figures, enabling fine-grained revision detection, localization, and structural interpretation.
\end{itemize}

\section{Related Work}
\label{relatedwork}

\subsection{Document comparison and differencing methods}

Existing studies on document comparison can be broadly grouped into two main paradigms. The first formulates comparison as a textual differencing problem, where different document versions are transformed into token sequences, lines, or discourse units, and revisions are identified through edit operations, sequence alignment, or semantic revision labeling \citep{shannon2010deep,yang2017identifying}. This paradigm has been widely adopted in revision tracking, editorial comparison, and change summarization, particularly in scenarios where document content can be represented under relatively stable textual ordering assumptions \citep{kuznetsov2022revise,ruan2024re3}. More recent work has further extended this line of research by incorporating semantic similarity modeling and neural revision classification, allowing document comparison to move beyond surface-level string edits toward higher-level textual correspondence \citep{d2024aries,jourdan2024casimir}.

The second paradigm treats document comparison as a visual discrepancy detection problem. In this setting, document pages are rendered as images and compared through page registration, pixel-level differencing, or region-based visual matching \citep{yenigalla2023similar,zhang2025dockylin}. Compared with purely text-driven pipelines, such methods are generally more tolerant to OCR or extraction errors and can directly capture page-level visual modifications. Recent advances have also introduced layout-aware visual representations to improve robustness under page reformatting, local displacement, and visual rearrangement \citep{jourdan2025identifying,xie2026vip}.

Despite their effectiveness in general-purpose document comparison, these two paradigms remain largely constrained by their underlying comparison spaces. Text-based methods are primarily designed for sequential revision analysis, whereas visual methods mainly target appearance-level discrepancy detection. Consequently, most existing approaches operate on a single modality of comparison and do not provide a unified mechanism for comparing semantically typed document components under heterogeneous representations. This limitation is particularly evident in scientific documents, where text, tables, formulas, and figures exhibit fundamentally different organizational principles and revision patterns, and therefore often require type-specific alignment and comparison strategies.

\subsection{Document intelligence for structured scientific documents}

In parallel with the development of document comparison methods, recent advances in document intelligence have substantially strengthened machine understanding of structured and visually rich documents. Representative research directions include page layout analysis, document object detection, table structure recognition, mathematical expression parsing, figure understanding, and multimodal document representation learning \citep{pfitzmann2022doclaynet,yu2024survey,dong2025mmdocir,guo2025vision}. Collectively, these studies have established effective foundations for identifying semantically meaningful regions and recovering the internal organization of complex document elements, especially in scientific and technical documents \citep{hu2025mplug,luo2026hainougat}.

This body of work is highly relevant to our setting because it enables type-aware representation learning for heterogeneous document content. In particular, layout analysis supports spatially grounded page decomposition \citep{pena2024continuous,shehzadi2025docsemi}, while table understanding preserves row--column topology and merged-cell relations \citep{wang2004table,kang2026benchmark}. At the same time, formula parsing captures symbolic and hierarchical structure beyond surface rendering \citep{kostalia2020evaluating,aggarwal2024survey}, and figure modeling provides region-level visual representations for scientific graphics \citep{wang2017graphs,das2023key}. Together, these advances make it possible to represent scientific documents as structured collections of semantic elements rather than as plain text streams or page images alone.

However, most existing document intelligence studies are centered on single-document understanding, with primary objectives such as detection, recognition, parsing, classification, or representation learning. Comparatively less attention has been devoted to cross-version reasoning over structured documents, where the problem is not limited to understanding isolated elements, but further requires establishing stable correspondence across versions and interpreting changes after alignment in a structure-aware manner. Therefore, although document intelligence provides essential technical foundations for structured document modeling, it does not by itself resolve the broader problem of cross-version differencing for scientific documents. This gap calls for a unified framework that integrates heterogeneous element decomposition, cross-version alignment, and structure-aware differencing.

\section{Problem Formulation}
\label{problemformulation}

Document differencing in scientific and technical publishing is substantially more challenging than conventional text comparison, because the compared objects are not homogeneous token sequences but page-structured documents composed of heterogeneous semantic elements, such as text blocks, tables, mathematical formulas, and figures. These elements differ in representation space, structural organization, and revision patterns, making traditional sequence-based diff or whole-page visual comparison insufficient for reliable cross-version analysis.

Given two document versions, denoted by $D^{(a)}$ and $D^{(b)}$, we model each document as a set of heterogeneous elements:

\begin{equation}
\mathcal{E}^{(a)} = \bigcup_{k \in \mathcal{K}} \mathcal{E}^{(a)}_k,\qquad
\mathcal{E}^{(b)} = \bigcup_{k \in \mathcal{K}} \mathcal{E}^{(b)}_k,
\end{equation}
where $\mathcal{K}=\{\text{text}, \text{table}, \text{formula}, \text{figure}\}$, and $\mathcal{E}^{(a)}_k$ denotes the set of elements of type $k$ extracted from $D^{(a)}$. Each element is represented as $ e = (b,c,s,t)$, where $b$ denotes the spatial region, $c$ the content representation, $s$ denotes the structure-aware representation, which may additionally encode typography-sensitive cues for text elements, and $t$ denotes the semantic type.

For each element type $k$, we define a cross-version alignment mapping $\mathcal{A}_k :\mathcal{E}^{(a)}_k \rightarrow \mathcal{E}^{(b)}_k \cup {\varnothing}$, where $\varnothing$ indicates that an element has no valid counterpart in the other version. Based on the aligned element pairs, the differencing objective is to predict

\begin{equation}
    \Delta_k\left(e_i^{(a)}, e_j^{(b)}\right)=\left(y^{det}_{ij},y^{loc}_{ij},y^{str}_{ij}\right),
\end{equation}
where $y^{det}_{ij}$ denotes whether the aligned pair has changed, $y^{loc}_{ij}$ denotes the localized changed region or sub-unit, and $y^{str}_{ij}$ denotes the structural comparison result.

Accordingly, complex document differencing is formulated as a unified problem of heterogeneous element decomposition, cross-version alignment, and structure-aware difference reasoning. This formulation differs from conventional document diff methods in that it operates on semantically typed elements rather than a single text sequence or page image, and it explicitly treats alignment as a prerequisite for reliable differencing.

\begin{algorithm}[t]
\SetAlCapFnt{\footnotesize}
\caption{\small{Unified inference for document differencing}}
\label{alg:unified_diff}
\footnotesize
\KwIn{Two versions of a scientific document, $D^{(a)}$ and $D^{(b)}$}
\KwOut{A structured set of differencing results, $\mathcal{Y}$}

\BlankLine
Initialize element sets and output: $\mathcal{E}^{(a)} \leftarrow \varnothing$, $\mathcal{E}^{(b)} \leftarrow \varnothing$,  $\mathcal{Y} \leftarrow \varnothing$\; 
\tcp*[h]{Heterogeneous element decomposition} \\
\ForEach{$(D,\mathcal{E}) \in \{(D^{(a)},\mathcal{E}^{(a)}), (D^{(b)},\mathcal{E}^{(b)})\}$}{
    \ForEach{page $p_i \in D$}{
        normalize $p_i$ and detect semantic regions $\mathcal{R}_i$\;
        \ForEach{region $r_{i,j} \in \mathcal{R}_i$}{
            construct typed element $e_{i,j}=(b,c,s,t)$ from $r_{i,j}$\;
            $\mathcal{E} \leftarrow \mathcal{E} \cup \{e_{i,j}\}$\;
        }
    }
}
\tcp*[h]{Cross-version element alignment} \\
\ForEach{$k \in \mathcal{K}$}{ 
    extract typed element sets $\mathcal{E}_k^{(a)}$ and $\mathcal{E}_k^{(b)}$\;
    \ForEach{$e_i^{(a)} \in \mathcal{E}_k^{(a)}$}{
        find a valid aligned match for $e_i^{(a)}$ in $\mathcal{E}_k^{(b)}$\;
        \eIf{no valid match is found}{
            set $\mathcal{A}_k(e_i^{(a)}) \leftarrow \varnothing$ and record insertion/deletion in $\mathcal{Y}$\;
        }{
            set $\mathcal{A}_k(e_i^{(a)})$ to the matched element\;
        }
    }
}
\tcp*[h]{Structure-aware differencing} \\
\ForEach{$k \in \mathcal{K}$}{
    \ForEach{$e_i^{(a)} \in \mathcal{E}_k^{(a)}$ such that $\mathcal{A}_k(e_i^{(a)}) \neq \varnothing$}{
        generate a structured differencing output for $(e_i^{(a)}, \mathcal{A}_k(e_i^{(a)}))$\;
        add the output to $\mathcal{Y}$\;
    }
}

\Return{$\mathcal{Y}$}\;
\end{algorithm}

\section{Methodology}
\label{methodology}

\subsection{Overview}

\begin{figure*}[t]
    \centering
    \includegraphics[width=1\linewidth]{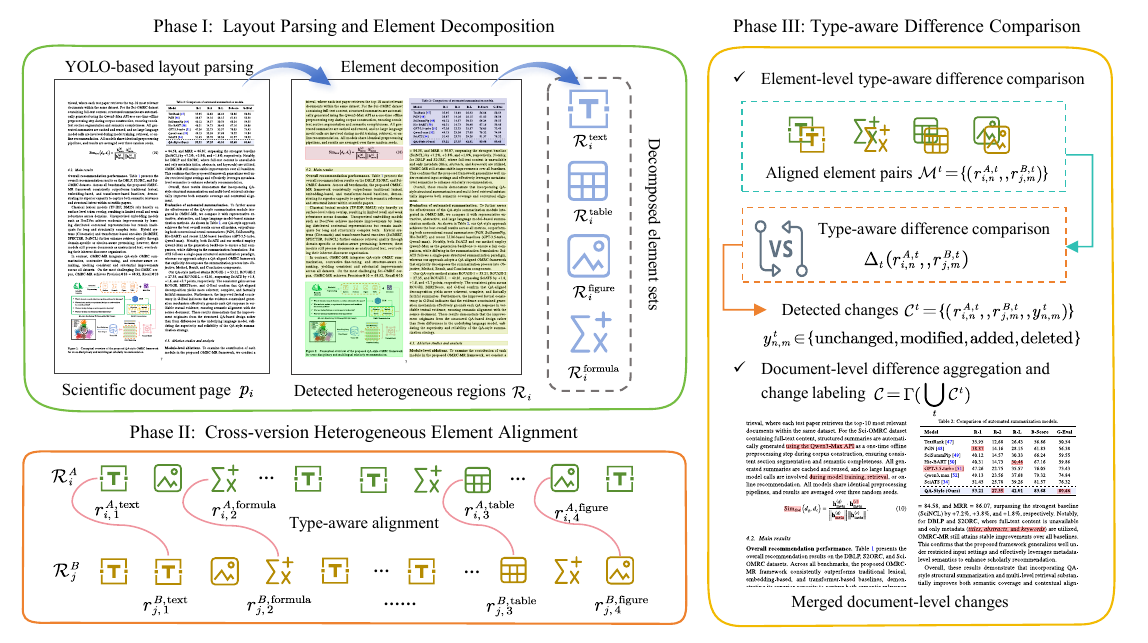}
\vspace{-.6cm}
\caption{Overview of the proposed heterogeneous element-aware document comparison framework. The framework consists of three phases: layout parsing and element decomposition, cross-version heterogeneous element alignment, and type-aware difference comparison with document-level change aggregation.}
\vspace{-.1cm}
\label{fig:Figure2}
\end{figure*}

To address the heterogeneous and structure-sensitive nature of complex document differencing, we propose a layout-aware and element-specific comparison framework consisting of three stages: heterogeneous element decomposition, cross-version element alignment, and structure-aware difference reasoning. Fig.~\ref{fig:Figure2} illustrates the overall architecture, and Algorithm~\ref{alg:unified_diff} summarizes the corresponding inference procedure.

Given two document versions, the framework first performs page normalization and layout analysis to decompose each document into semantically typed elements, including text blocks, tables, formulas, and figures. Different from conventional diff methods that directly compare linearized text or whole-page images, this stage provides explicit spatial anchors and type-aware comparison units. Each detected element is then converted into a structured representation tailored to its semantic type, so that textual content together with typography-sensitive cues, tabular organization, mathematical structure, and visual regions can be compared in their appropriate representation spaces.

In the second stage, the framework establishes cross-version correspondence between heterogeneous elements before differencing. This alignment-first design is critical for scientific documents, where revisions often involve paragraph movement, table reformatting, formula relocation, and figure rearrangement. Instead of assuming fixed sequential correspondence, the proposed method aligns candidate elements by jointly modeling layout proximity, content similarity, and structural compatibility, thereby reducing spurious differences caused by reflow or local reorganization.

In the third stage, the framework performs structure-aware difference reasoning over aligned element pairs. For each pair, it predicts whether a change has occurred, localizes the changed region or sub-unit, and analyzes whether the internal structural relations remain consistent across versions. The overall pipeline is therefore layout-aware, element-specific, and alignment-first with structure-aware reasoning. Through these design choices, the proposed framework provides a unified solution for cross-version differencing of scientific documents while preserving the intrinsic heterogeneity of text, tables, formulas, and figures.

\subsection{Layout-aware heterogeneous element decomposition}

The first stage transforms each document page into a set of semantically typed and spatially grounded elements. This step is essential because complex document differencing cannot be reliably performed on linearized text or whole-page images alone; it requires explicit element boundaries, semantic categories, and type-appropriate representations.

Given an input document version, each page is first normalized through rendering, coordinate calibration, and orientation correction, so that different document versions can be compared under a unified spatial reference system. A lightweight YOLO-based layout detector, fine-tuned for scientific document pages, is then applied to identify heterogeneous semantic regions on each page. For page $p_i$, the detector outputs

\begin{equation}
    \mathcal{R}_i = \left\{r_{i,1}, r_{i,2}, \dots, r_{i,m_i}\right\},
\end{equation}
where each region $r_{i,j}$ is associated with a bounding box, a semantic label, and a confidence score. In this work, the primary semantic categories include text, table, formula, and figure.

For each detected region, we instantiate the element-specific representation introduced in Section~\ref{problemformulation}, so that heterogeneous document elements can be compared in type-appropriate content and structure spaces rather than under a single low-level representation.

For text elements, textual content is extracted together with character- or span-level typographic attributes, from which paragraph/sentence organization and typography-sensitive cues are recovered, including font family, font size, bold/italic style, and superscript/subscript information. For table elements, the detected region is structurally parsed into a cell-based representation and further converted into an HTML tree or a grid-based matrix, preserving row–column organization and merged-cell relations. For formula elements, the detected region is transformed into a symbolic or structure-aware representation, such as LaTeX, MathML, or an operator tree, so that mathematical relations are explicitly preserved. For figure elements, we retain the cropped visual region and extract local feature representations, optionally augmented with caption or subfigure associations when available.

Through this stage, each document is converted into a structured collection of heterogeneous semantic elements with explicit spatial grounding and type-aware content and structure representations, providing the basis for subsequent alignment and difference reasoning.

\subsection{Cross-version heterogeneous element alignment}

To establish stable correspondence across document revisions, the proposed framework performs cross-version alignment before element-wise differencing. Since revisions in scientific documents may involve local reordering, layout reflow, or element relocation, correspondence cannot be reliably inferred from sequential order or absolute page position alone. We therefore align heterogeneous elements in a type-specific manner by jointly considering their layout, content, and structural compatibility.

For each element type $k \in \mathcal{K}$, the aligned counterpart of an element $e_i^{(a)} \in \mathcal{E}_k^{(a)}$ is selected from the candidate set $\mathcal{E}_k^{(b)}$ according to a type-specific compatibility score. The alignment score between two candidate elements $e_i^{(a)}$ and $e_j^{(b)}$ is defined as

\begin{equation}
\begin{aligned}
\Phi_k\left(e_i^{(a)}, e_j^{(b)}\right) 
&= \lambda_{k,1} \cdot \phi_k^{loc}\left(e_i^{(a)}, e_j^{(b)}\right)
+ \lambda_{k,2} \cdot \phi_k^{cont}\left(e_i^{(a)}, e_j^{(b)}\right)  \\
&\quad + \lambda_{k,3} \cdot \phi_k^{str}\left(e_i^{(a)}, e_j^{(b)}\right).
\end{aligned}
\label{eq4}
\end{equation}
where $\phi_k^{loc}$, $\phi_k^{cont}$, and $\phi_k^{str}$ denote the location, content, and structural compatibility terms, respectively, and $\lambda_{k,1}$, $\lambda_{k,2}$, and $\lambda_{k,3}$ are type-specific balance coefficients for element type $k$. Accordingly, the aligned counterpart of $e_i^{(a)}$ is obtained as

\begin{equation}
\mathcal{A}_k\left(e_i^{(a)}\right)=
\arg\max_{e_j^{(b)} \in \,\mathcal{E}^{(b)}_k}
\Phi_k\left(e_i^{(a)}, e_j^{(b)}\right).
\end{equation}

To ensure alignment reliability, the resulting assignment is retained only if the maximum compatibility score is no lower than the type-specific alignment confidence threshold $\tau_k$; otherwise, the element is regarded as unmatched and handled as an insertion or deletion case in subsequent differencing.

The location term $\phi_k^{loc}$ measures spatial consistency under the normalized page coordinate system. The content term $\phi_k^{cont}$ evaluates similarity in the corresponding content space, such as textual semantics, tabular content, symbolic formula representation, or visual features. The structural term $\phi_k^{str}$ captures compatibility in internal organization, including paragraph structure, table topology, mathematical expression hierarchy, or figure layout.

Although the alignment formulation is unified, its instantiation remains element-specific. For text elements, alignment is mainly performed at the paragraph or text-block level, with optional sentence-level refinement within matched blocks. For table elements, alignment is first established at the table level and can be further refined using cell-, row-, or column-level structural consistency. For formula elements, symbolic and structural compatibility play a more important role than raw visual similarity. For figure elements, alignment combines region-level visual similarity with layout proximity and, when available, caption or internal-region associations.

By explicitly establishing correspondence before comparison, this stage reduces false differences caused by element displacement, pagination changes, and layout restructuring, thereby providing stable comparison units for downstream localization and structure-aware reasoning.

\subsection{Structure-Aware Difference Reasoning}

After cross-version alignment, the framework performs structure-aware difference reasoning over each matched element pair. Rather than relying on coarse binary comparison alone, this stage jointly analyzes whether a revision has occurred, where the change is localized, and whether the internal structure remains consistent when such structural information is meaningful for the corresponding element type. For each aligned pair, the framework produces a structured differencing output that reflects change detection, change localization, and, where applicable, structural consistency.

For text elements, the framework compares aligned text blocks in a coarse-to-fine manner, localizes revisions at the sentence or span level, and further captures typography-sensitive changes such as font substitution, font-size variation, bold/italic shifts, and superscript/subscript modifications. It also considers structural adjustments in heading association, paragraph ordering, and local textual organization. For table elements, the reasoning jointly examines cell content and structural topology, enabling cell-level localization as well as the identification of row/column adjustments, header modifications, and merged-cell changes, thereby distinguishing content edits from structural reorganization within a table. For formula elements, the reasoning emphasizes symbolic and structural consistency rather than appearance alone, focusing on operator hierarchy, superscript/subscript dependencies, and expression-level relations so as to distinguish genuine mathematical revisions from superficial rendering variation. For figure elements, the framework focuses on region-level visual differencing by detecting revised visual areas and evaluating their spatial correspondence across versions, which is suitable for scientific figures where local visual regions may be updated, replaced, or relocated.

By performing type-aware reasoning on aligned heterogeneous elements, this stage extends document differencing from coarse change detection to interpretable multi-level analysis. It supports fine-grained localization and structural interpretation for elements with explicit internal structure, while retaining region-level visual reasoning for figure elements whose large-scale component-level structural annotations are unavailable.

\subsection{Unified inference framework and output formulation}

The proposed framework integrates the preceding stages into a unified heterogeneous differencing pipeline. Given two document versions $D^{(a)}$ and $D^{(b)}$, it first performs layout-aware heterogeneous element decomposition on each document, then establishes cross-version correspondence for each element type through the alignment mapping $\mathcal{A}_k$, and finally applies structure-aware difference reasoning to each aligned element pair. As a result, document differencing is performed on explicitly aligned heterogeneous comparison units rather than on linearized text sequences or whole-page images.

For each element type $k \in \mathcal{K}$, the final document-level differencing result is expressed as

\begin{equation}
    \mathcal{Y} = \bigcup_{k \in \mathcal{K}}
\left\{
\Delta_k\left(e_i^{(a)}, \mathcal{A}_k(e_i^{(a)})\right)
\mid e_i^{(a)} \in \mathcal{E}^{(a)}_k
\right\},
\end{equation}
where $\mathcal{Y}$ denotes the complete set of structured difference outputs over all heterogeneous element types. Each output item records whether a change is detected, where the change is localized, and whether the internal structure remains consistent across versions.

A key property of the proposed framework is that it is heterogeneous in representation but unified in output semantics. Although text, tables, formulas, and figures are represented and compared in different type-appropriate spaces, their reasoning results are projected onto a common output space characterized by four evaluation-oriented dimensions: change detection, change localization, structure awareness, and alignment/matching quality. This unified output formulation provides the basis for the evaluation protocol used in the subsequent experimental section.

\section{Experiment and Analysis}

\subsection{Experimental setup}
\noindent
\textbf{Dataset.}
The experimental dataset was constructed from real-world document comparison data generated in the production proofreading workflows of journal editorial offices served by our enterprise system. Each sample consists of two scientific PDF versions from adjacent or related proofreading stages, such as typesetting, author correction, editorial proofreading, and pre-publication verification. Unlike synthetic revision datasets, these document pairs reflect practical cross-version changes that occur in real scholarly publishing environments. The dataset covers journals from multiple disciplines and includes diverse page layouts, such as single-column and multi-column formats, formula-intensive pages, complex tables, figures with captions, and heterogeneous typography conventions.

For evaluation, each document pair is decomposed into heterogeneous element-level comparison units, including text, tables, formulas, and figures. The resulting annotations support the evaluation of element alignment, change detection, change localization, and structure-aware differencing. These characteristics make the dataset suitable for assessing document differencing methods under realistic layout variation, element relocation, table restructuring, formula modification, figure replacement, and typography-sensitive revision scenarios. Detailed dataset construction procedures, statistics, discipline distribution, proofreading-stage distribution, element-type distribution, and annotation protocol are reported in Appendix~A.

\noindent
\textbf{Baseline methods.}
To ensure a fair and element-aware comparison, the baselines are grouped according to the four semantic element types considered in this work. For \textit{text elements}, we compare with TextDiff \cite{ruan2024re3}, which performs standard edit-script-based differencing over extracted text sequences, SentenceDiff \cite{d2024aries}, which aligns and compares sentence-level units, SemanticTextDiff \cite{lan2025unit}, which uses semantic similarity between text embeddings, and TypographyDiff \cite{shi2025fonts}, which explicitly compares font family, font size, bold/italic style, and superscript/subscript attributes. For \textit{table elements}, we include TableTextDiff \cite{nunes2025benchmarking}, which linearizes table content for textual comparison, CellDiff \cite{kudale2024sprint}, which performs cell-level matching after row--column reconstruction, and TableStructDiff \cite{yang2025enhanced}, which compares grid-, HTML-, or tree-based table representations to capture row/column adjustments, header modifications, and merged-cell changes.

For \textit{formula elements}, we evaluate LaTeXDiff \cite{horn2025benchmarking}, which compares LaTeX token sequences, SymbolDiff \cite{wieckowiak2025multimodal}, which matches recognized mathematical symbols and operators, and FormulaTreeDiff \cite{zhu2025tamer}, which compares MathML or operator-tree representations using structural similarity. For \textit{figure elements}, we compare with ImageDiff \cite{li2025document, jain2013visualdiff}, which detects pixel-level discrepancies, SSIMDiff \cite{wang2004image, wickrema2025benchmarking}, which measures structural image similarity, PerceptualDiff \cite{zhang2018unreasonable}, which compares deep visual features, and RegionDiff \cite{zhang2025class}, which performs local region matching within cropped figure areas. 

\noindent
\textbf{Implementation details.}
All document pairs are processed using a unified preprocessing pipeline. Each PDF page is rendered at 200 dpi for page-level layout parsing, while cropped formula and figure regions are re-rendered at 300 dpi to preserve fine-grained visual details. Page coordinates are normalized to $[0,1]$ after page-size calibration and orientation correction. Heterogeneous document regions are detected using a YOLOv10m-based layout detector fine-tuned on scientific document pages. The detector takes images resized to $1280 \times 1280$ as input, uses a confidence threshold of 0.25, and keeps at most 300 region proposals per page. Following the end-to-end detection design of YOLOv10, no additional NMS post-processing is applied during inference. For each detected region, a type-specific representation is constructed. Text elements are extracted from the native PDF text layer using PyMuPDF and represented by textual content together with span-level typographic attributes, including font family, font size, bold/italic style, and superscript/subscript status. OCR is used only as a fallback for pages without reliable embedded text. Table elements are represented by reconstructed cell grids and, when structural parsing succeeds, HTML-style structures generated by a PP-StructureV2/SLANet-style parser. Formula elements are represented by LaTeX sequences, recognized symbols, or MathML/operator-tree structures produced by an offline formula parser. Figure elements are represented by cropped visual regions, DINOv2 ViT-B/14 features, and LPIPS-style perceptual features. All bounding boxes are stored in normalized page coordinates to support consistent spatial comparison across document versions.

For the proposed method, cross-version matching is restricted to elements of the same semantic type. Candidate retrieval is first performed within a soft page window of $\pm 2$ pages and is further supplemented by content- or visual-similarity top-$K$ retrieval, with $K=50$, to handle large element displacement caused by page reflow or local reorganization. The alignment score combines location, content, and structural cues, with type-specific balance coefficients selected on the validation set and fixed during testing. We set $(\lambda_{k,1},\lambda_{k,2},\lambda_{k,3})$ to $(0.20,0.60,0.20)$ for text, $(0.15,0.35,0.50)$ for tables, $(0.10,0.35,0.55)$ for formulas, and $(0.25,0.60,0.15)$ for figures. The corresponding alignment confidence thresholds for text, table, formula, and figure elements are set to 0.72, 0.66, 0.70, and 0.64, respectively. Elements whose best matching scores fall below the corresponding thresholds are treated as unmatched and handled as insertion or deletion cases. All baseline methods are evaluated on the same document pairs, data splits, and automatically detected element proposals as the proposed method. The text, table, formula, and figure baselines are implemented using their respective comparison signals, including edit-script matching, sentence or semantic similarity, typography comparison, cell/grid/tree matching, symbolic or structural formula comparison, and pixel-, SSIM-, perceptual-, or region-level visual comparison. During testing, ground-truth element correspondence is not provided to any method; thus, all methods are evaluated under an end-to-end cross-version differencing setting. Key implementation settings are summarized in Table~\ref{tab:key_implementation_settings}, and full software versions, hardware configuration, detector fine-tuning details, and complete hyperparameter settings are provided in Appendix~B.

\begin{table}[t]
\centering
\begin{threeparttable}
\caption{Key implementation settings used in the main experiments.}
\label{tab:key_implementation_settings}
\footnotesize
\setlength{\tabcolsep}{3pt}
\renewcommand{\arraystretch}{1.12}
\begin{tabular}{@{}p{0.45\columnwidth}p{0.50\columnwidth}@{}}
\toprule
\textbf{\; Item} & \textbf{Setting} \\
\midrule
\, Page rendering resolution & 200 dpi \\
\, Formula/figure crop rendering & 300 dpi \\
\, Layout detector & YOLOv10m-based detector \\
\, Detector input size & $1280 \times 1280$ \\
\, Detection confidence threshold & 0.25 \\
\, Maximum detections per page & 300 \\
\, Detector post-processing & NMS-free inference \\
\, Candidate page window & $\pm 2$ pages \\
\, Top-$K$ retrieval candidates & 50 \\
\, Text extractor & PyMuPDF with OCR fallback \\
\, Text semantic encoder & BGE-M3 \\
\, Table parser & PP-StructureV2\allowbreak/SLANet-style parser \\
\, Formula parser & Offline LaTeX\allowbreak/MathML parser \\
\, Figure feature extractor & DINOv2 ViT-B/14 and LPIPS-style features \\
\, Alignment thresholds & Text/Table/Formula/Figure: 0.72/0.66/0.70/0.64 \\
\, Test-time correspondence & No ground-truth correspondence \\
\bottomrule
\end{tabular}
\end{threeparttable}
\end{table}

\noindent
\textbf{Evaluation metrics.}
We evaluate document differencing performance from four complementary dimensions: change detection, change localization, structure awareness, and alignment/matching quality. All metrics are computed at the element level and reported separately for text, table, formula, and figure elements. For \textit{change detection}, we report Precision, Recall, and F1-score to measure whether a method correctly identifies changed elements, including insertion and deletion cases. For \textit{change localization}, we report localization F1 and IoU. Localization F1 is computed over type-specific atomic units, including text spans or sentences, table cells, formula tokens or subtrees, and figure subregions, while IoU is used for spatially localized changes. For \textit{structure awareness}, we report Structure Similarity and Structure Consistency to evaluate whether a method can preserve or identify changes in internal organization, such as paragraph structure, table topology, formula hierarchy, and figure layout relations.

For \textit{alignment and matching quality}, we report Alignment Accuracy and Matching F1. Alignment Accuracy measures whether elements with valid cross-version counterparts are correctly aligned, whereas Matching F1 evaluates the predicted matched-pair set against the ground-truth correspondence set, including the ability to distinguish matched and unmatched elements. Unless otherwise specified, all scores are first averaged over document pairs and then macro-averaged across element types to avoid domination by frequent text elements. 
All baseline outputs are projected into the same evaluation space through deterministic post-processing based on their predicted matches, changed units, and recovered element attributes. Therefore, a metric is reported whenever it is semantically meaningful for the corresponding element type and can be derived from the method output. N/A is used only when the metric is not applicable to the element type or annotation setting, and such entries are excluded from macro-averaging.
Detailed mathematical definitions of all metrics, type-specific localization units, structural similarity functions, and averaging rules are provided in Appendix~C.

\subsection{Main results}

\begin{table*}[t]
\centering
\begin{threeparttable}
\caption{Main comparison results of the proposed framework and element-specific baselines across semantic element types and evaluation dimensions. The best result within each element group is highlighted in bold.}
\label{tab:main_results}
\footnotesize 
\setlength{\tabcolsep}{6.8pt}
\renewcommand{\arraystretch}{1.1}
\begin{tabular}{llccccccccc}
\toprule
\multirow{2}{*}{Element} 
& \multirow{2}{*}{Method}
& \multicolumn{3}{c}{Change Detection}
& \multicolumn{2}{c}{\; Change Localization \;}
& \multicolumn{2}{c}{\; Structure Awareness \;}
& \multicolumn{2}{c}{\; Alignment / Matching \;} \\
\cmidrule(lr){3-5}
\cmidrule(lr){6-7}
\cmidrule(lr){8-9}
\cmidrule(lr){10-11}
& & \; Precision & Recall & Det-F1 \; & \; Loc-F1 & IoU & \; Str-Sim & Str-Cons & \; Align-Acc & Match-F1 \\
\midrule

\multirow{5}{*}{Text}
& TextDiff 
& 0.823 & 0.781 & 0.801 & 0.742 & N/A & 0.512 & 0.628 & 0.762 & 0.777 \\
& SentenceDiff 
& 0.836 & 0.809 & 0.822 & 0.789 & N/A & 0.548 & 0.654 & 0.807 & 0.815 \\
& SemanticTextDiff \; 
& 0.817 & 0.842 & 0.829 & 0.765 & N/A & 0.536 & 0.648 & 0.828 & 0.829 \\
& TypographyDiff 
& 0.861 & 0.798 & 0.828 & 0.802 & N/A & 0.731 & 0.782 & 0.812 & 0.820 \\
& \textbf{Ours} 
& \textbf{0.913} & \textbf{0.894} & \textbf{0.903} & \textbf{0.871} & N/A & \textbf{0.842} & \textbf{0.879} & \textbf{0.914} & \textbf{0.908} \\
\midrule

\multirow{4}{*}{Table}
& TableTextDiff 
& 0.702 & 0.651 & 0.676 & 0.598 & 0.556 & 0.421 & 0.536 & 0.628 & 0.643 \\
& CellDiff 
& 0.761 & 0.718 & 0.739 & 0.701 & 0.667 & 0.587 & 0.642 & 0.705 & 0.716 \\
& TableStructDiff 
& 0.789 & 0.752 & 0.770 & 0.724 & 0.698 & 0.716 & 0.758 & 0.741 & 0.752 \\
& \textbf{Ours} 
& \textbf{0.872} & \textbf{0.839} & \textbf{0.855} & \textbf{0.804} & \textbf{0.776} & \textbf{0.833} & \textbf{0.861} & \textbf{0.846} & \textbf{0.852} \\
\midrule

\multirow{4}{*}{Formula}
& LaTeXDiff 
& 0.754 & 0.703 & 0.728 & 0.682 & N/A & 0.655 & 0.688 & 0.700 & 0.709 \\
& SymbolDiff 
& 0.781 & 0.735 & 0.757 & 0.716 & N/A & 0.674 & 0.704 & 0.724 & 0.733 \\
& FormulaTreeDiff 
& 0.806 & 0.761 & 0.783 & 0.748 & N/A & 0.769 & 0.797 & 0.759 & 0.767 \\
& \textbf{Ours} 
& \textbf{0.878} & \textbf{0.846} & \textbf{0.862} & \textbf{0.823} & N/A & \textbf{0.852} & \textbf{0.877} & \textbf{0.835} & \textbf{0.842} \\
\midrule

\multirow{5}{*}{Figure}
& ImageDiff 
& 0.668 & 0.724 & 0.695 & N/A & 0.503 & N/A & N/A & 0.604 & 0.621 \\
& SSIMDiff 
& 0.701 & 0.693 & 0.697 & N/A & 0.538 & N/A & N/A & 0.642 & 0.655 \\
& PerceptualDiff 
& 0.762 & 0.718 & 0.739 & N/A & 0.591 & N/A & N/A & 0.698 & 0.704 \\
& RegionDiff 
& 0.789 & 0.742 & 0.765 & N/A & 0.646 & N/A & N/A & 0.744 & 0.751 \\
& \textbf{Ours} 
& \textbf{0.862} & \textbf{0.829} & \textbf{0.845} & N/A & \textbf{0.721} & N/A & N/A & \textbf{0.828} & \textbf{0.835} \\

\bottomrule
\end{tabular}
\begin{tablenotes}[flushleft]
\footnotesize
\item \textbf{N/A} denotes metrics that are not applicable to the element type or annotation setting and is excluded from macro-averaging. IoU is not reported for text and formula elements because their localization is evaluated over discrete textual or symbolic units. For figure elements, Loc-F1 and structure-awareness metrics are omitted because large-scale component-level structural annotations are unavailable.
\end{tablenotes}
\end{threeparttable}
\vspace{-.25cm}
\end{table*}

\noindent
\textbf{Overall performance.}
Table~\ref{tab:main_results} presents the main comparison results across heterogeneous document elements and grouped evaluation dimensions. All methods are evaluated under the same element-level protocol, and baseline outputs are mapped into the common evaluation space through deterministic post-processing when the corresponding metric is meaningful. N/A denotes metrics that are not applicable to the element type or annotation setting and is excluded from macro-averaging. Specifically, IoU is not reported for text and formula elements because their localization is evaluated over sentence/span units or symbolic substructures rather than spatial regions. For figure elements, localization is evaluated by region-level IoU, while Loc-F1 and structure-awareness metrics are omitted because large-scale component-level structural annotations are unavailable in the production dataset.

Overall, the proposed method consistently achieves the best performance across all applicable metrics and all four element types. For text elements, our method obtains a detection F1 of 0.903, outperforming the strongest text baseline, SemanticTextDiff, by 7.4 percentage points. It also improves Loc-F1 from 0.802 to 0.871 compared with TypographyDiff, and achieves the highest structure-awareness and matching scores. These results indicate that combining textual content, layout correspondence, and typography-sensitive attributes is more effective than relying solely on sequence-level, sentence-level, semantic, or typography-only comparison.

For table and formula elements, the advantage of structure-aware differencing is more pronounced. On tables, TableStructDiff performs better than TableTextDiff and CellDiff because it preserves grid and structural information, but it still lags behind the proposed method across detection, localization, structure awareness, and alignment metrics. Our method improves table detection F1 from 0.770 to 0.855, structure similarity from 0.716 to 0.833, and matching F1 from 0.752 to 0.852 over the strongest table baseline. For formulas, FormulaTreeDiff provides the strongest baseline by modeling symbolic hierarchy, yet our method further improves detection F1 from 0.783 to 0.862 and structure consistency from 0.797 to 0.877. This confirms that element-specific alignment combined with structural reasoning is critical for tables and mathematical expressions, where row--column topology, merged-cell relations, operator hierarchy, and superscript/subscript dependencies directly affect semantic interpretation.

For figure elements, visual baselines show a gradual improvement from ImageDiff and SSIMDiff to PerceptualDiff and RegionDiff, reflecting the benefit of perceptual features and local region matching. Nevertheless, these methods mainly operate in the visual space and remain sensitive to rendering variation, local displacement, and figure relocation. The proposed method achieves a detection F1 of 0.845, IoU of 0.721, alignment accuracy of 0.828, and matching F1 of 0.835, outperforming RegionDiff by 8.0, 7.5, 8.4, and 8.4 percentage points, respectively. These gains demonstrate that cross-version element alignment and type-aware region reasoning are important even for visually dominated elements.

Taken together, the results show that no single baseline strategy is sufficient for all scientific document elements. Textual methods are effective for content edits but weak in layout and structure preservation; table and formula baselines benefit from structured representations but still suffer from unstable cross-version correspondence; and visual baselines can capture figure-level changes but lack robust semantic alignment. By integrating heterogeneous element decomposition, type-specific alignment, and structure-aware differencing, the proposed framework provides a more robust and interpretable solution for cross-version scientific document differencing.

\noindent
\textbf{Effect of type-aware reasoning.}
To examine how the proposed framework handles typical revision patterns in production proofreading scenarios, we evaluate representative type-sensitive revision subsets in Table~\ref{tab:type_sensitive_subsets}. Each subset is measured using the most relevant metric from the main evaluation protocol. The proposed method achieves consistent gains over the strongest baseline across all subsets. For text elements, the improvement is moderate on ordinary content edits, increasing Det-F1 from 0.887 to 0.926, but becomes more pronounced on typography edits, where Str-Cons improves from 0.814 to 0.891. This indicates that font style, size, and superscript/subscript cues are meaningful revision signals in scientific proofreading.

For tables, the gain is larger on topology changes than on cell edits, with Str-Cons improving from 0.754 to 0.848, confirming the importance of row--column organization, header relations, and merged-cell structures. For formulas, hierarchy changes obtain a larger improvement than symbol changes, showing that operator hierarchy and superscript/subscript dependencies require structural reasoning beyond token-level comparison. For figures, the proposed method improves Det-F1 on figure replacement from 0.835 to 0.890 and IoU on region update from 0.631 to 0.704, demonstrating the benefit of combining visual-region reasoning with cross-version element alignment. These diagnostic results further support the necessity of type-aware differencing over heterogeneous scientific document elements.

\begin{table}[t]
\centering
\begin{threeparttable}
\caption{Diagnostic results on element-level type-sensitive revision subsets. Each subset is evaluated using the most relevant metric from the main evaluation protocol.}
\label{tab:type_sensitive_subsets}
\footnotesize
\setlength{\tabcolsep}{4.5pt}
\renewcommand{\arraystretch}{1.15}
\begin{tabular}{lllccc}
\toprule
\textbf{Element} & \textbf{Revision subset} & \textbf{Metric} & \textbf{Best base.} & \textbf{\; Ours \;} & \textbf{Gain} \\
\midrule
\multirow{2}{*}{Text} 
& Content edits & Det-F1 & 0.887 & \textbf{0.926} & +3.9 \\
& Typography edits & Str-Cons & 0.814 & \textbf{0.891} & +7.7 \\
\midrule
\multirow{2}{*}{Table} 
& Cell edits & Loc-F1 & 0.783 & \textbf{0.858} & +7.5 \\
& Topology changes & Str-Cons & 0.754 & \textbf{0.848} & +9.4 \\
\midrule
\multirow{2}{*}{Formula \;} 
& Symbol changes & Loc-F1 & 0.792 & \textbf{0.858} & +6.6 \\
& Hierarchy changes & Str-Cons & 0.789 & \textbf{0.861} & +7.2 \\
\midrule
\multirow{2}{*}{Figure} 
& Figure replacement & Det-F1 & 0.835 & \textbf{0.890} & +5.5 \\
& Region update & IoU & 0.631 & \textbf{0.704} & +7.3 \\
\bottomrule
\end{tabular}

\begin{tablenotes}[flushleft]
\footnotesize
\item \textbf{Best base.} denotes the strongest baseline selected within each revision subset. \textbf{Gains} are reported in percentage points.
\end{tablenotes}
\end{threeparttable}
\end{table}

\noindent
\textbf{Robustness under production proofreading scenarios.}
Different from the element-level type-sensitive subset analysis above, this analysis assigns each document pair to its dominant production proofreading scenario and evaluates performance at the document-pair level. As shown in Table~\ref{tab:proofreading_scenarios}, the proposed method consistently outperforms the strongest baseline across all scenarios, indicating that its advantage is not limited to a specific element type or evaluation dimension.

The gains are particularly clear in layout- and structure-sensitive scenarios. For text reflow, the proposed method improves Match-F1 from 0.742 to 0.826, demonstrating the benefit of cross-version alignment when paragraph positions change due to pagination or local layout adjustment. For table reformatting, Str-Cons increases from 0.746 to 0.842, confirming that row--column topology, header organization, and merged-cell relations cannot be reliably handled by content-only or cell-only comparison. Similarly, for formula hierarchy revisions, the proposed method improves Str-Cons from 0.782 to 0.854, showing the importance of symbolic and structural reasoning for mathematical expressions.

The proposed method also performs robustly in visually and typography-sensitive scenarios. For typography edits, Str-Cons improves from 0.804 to 0.886, indicating that font size, bold/italic style, and superscript/subscript cues are meaningful revision signals in scientific proofreading. For figure replacement, Det-F1 increases from 0.836 to 0.892, while for local region updates, IoU improves from 0.624 to 0.697. These results show that the framework can handle both global figure substitution and local visual modification. Overall, the scenario-level results provide independent evidence that the proposed alignment-first and type-aware differencing strategy is suitable for the revision patterns encountered in real production proofreading workflows.

\begin{table}[t]
\centering
\begin{threeparttable}
\caption{Robustness under representative production proofreading scenarios. Each document pair is assigned to its dominant proofreading scenario and evaluated using the most relevant metric from the main evaluation protocol.}
\label{tab:proofreading_scenarios}
\footnotesize
\setlength{\tabcolsep}{4pt}
\renewcommand{\arraystretch}{1.15}
\begin{tabular}{lclccc}
\toprule
\textbf{Scenario} & \textbf{Prop.} & \quad \textbf{Metric} & \textbf{Best base.} & \textbf{\, Ours \;} & \textbf{Gain} \\
\midrule
Text edit             & 24.1\% & \;\;\; Det-F1   & 0.881 & \textbf{0.918} & +3.7 \\
Typography edit       & 12.8\% & \;\;\; Str-Cons & 0.804 & \textbf{0.886} & +8.2 \\
Text reflow           & 16.5\% & \;\;\; Match-F1 & 0.742 & \textbf{0.826} & +8.4 \\
Table reformat        & 14.2\% & \;\;\; Str-Cons & 0.746 & \textbf{0.842} & +9.6 \\
Formula hierarchy     & 13.6\% & \;\;\; Str-Cons & 0.782 & \textbf{0.854} & +7.2 \\
Figure replacement \; & 11.2\% & \;\;\; Det-F1   & 0.836 & \textbf{0.892} & +5.6 \\
Region update         & 7.6\%  & \;\;\; IoU      & 0.624 & \textbf{0.697} & +7.3 \\
\bottomrule
\end{tabular}

\vspace{0.05mm}
\begin{tablenotes}[flushleft]
\footnotesize
\item \textbf{Prop.} denotes the proportion of each dominant proofreading scenario in the test set. \textbf{Best base.} denotes the strongest baseline under the corresponding scenario. \textbf{Gains} are reported in percentage points.
\end{tablenotes}
\end{threeparttable}
\end{table}

\begin{figure}[t]
    \centering
    \includegraphics[width=1\linewidth]{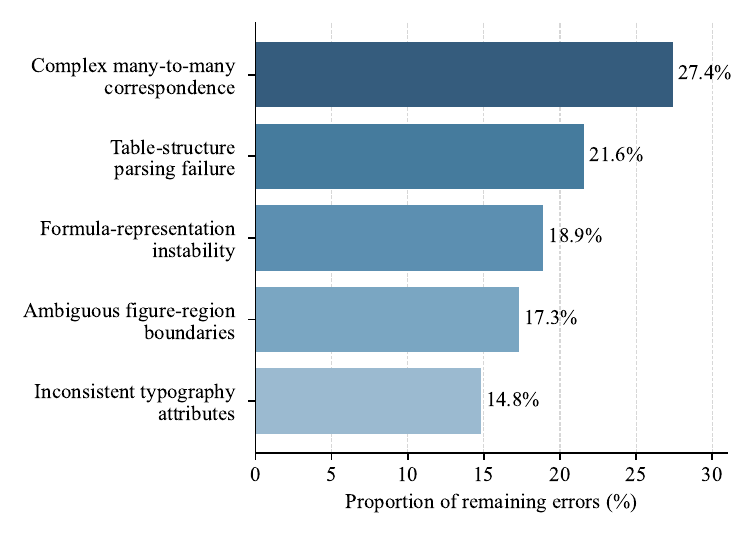}
\vspace{-.9cm}
\caption{Distribution of remaining error patterns on the test set. Proportions are computed over erroneous cases.}
\label{fig:Figure3}
\end{figure}

\noindent
\textbf{Remaining error patterns.}
To further examine the limitations of the proposed framework, we analyze the remaining erroneous cases on the test set and summarize the major error sources in Fig.~\ref{fig:Figure3}. The largest proportion of errors comes from complex many-to-many correspondence, accounting for 27.4\% of the remaining cases. Such errors typically occur when one paragraph is split into multiple blocks, several short paragraphs are merged, or tables are reorganized across pages. These cases go beyond simple one-to-one element matching and may lead to incorrect correspondence even when the changed content is partially detected.

Other errors mainly arise from imperfect upstream parsing and ambiguous local boundaries. Table-structure parsing failures account for 21.6\% of the remaining errors, often caused by dense grid lines, irregular merged cells, incomplete borders, or compact table layouts. Formula-representation instability contributes 18.9\%, especially when visually similar expressions produce inconsistent LaTeX, MathML, or operator-tree representations. Figure-region boundary ambiguity accounts for 17.3\%, reflecting the difficulty of precisely localizing subtle changes in curves, legends, markers, or embedded subregions. Typography attribute inconsistency accounts for 14.8\%, mainly due to incomplete or inconsistent font metadata in PDFs and the difficulty of distinguishing superscript/subscript status from local font-size variation. These findings indicate that future improvements should focus on more flexible many-to-many alignment, stronger table and formula parsers, and richer component-level annotations for scientific figures.

\subsection{Ablation studies and analysis}

\begin{figure}[t]
    \centering
    \includegraphics[width=1\linewidth]{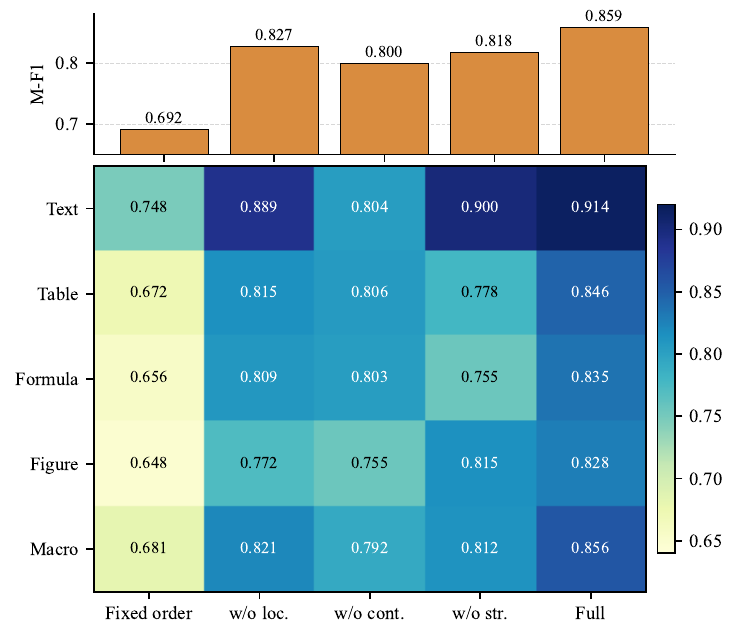}
\vspace{-.6cm}
\caption{Ablation analysis of cross-version alignment. (a) Matching F1 under different alignment variants. (b) Alignment accuracy across element types under different alignment variants.}
\label{fig:Figure4}
\end{figure}

\noindent
\textbf{Core module-level ablation.}
Table~\ref{tab:ablation_core} reports the contribution of the main components in the proposed framework. Removing cross-version alignment leads to the largest degradation, reducing D-F1 from 0.866 to 0.779, L-F1 from 0.833 to 0.706, A-Acc from 0.856 to 0.681, and M-F1 from 0.859 to 0.692. This variant degenerates the proposed alignment module into fixed page-order or reading-order matching, which corresponds to the \textit{Fixed order} setting further analyzed in Fig.~\ref{fig:Figure4}. This confirms that stable element correspondence is a prerequisite for reliable cross-version differencing; without explicit alignment, downstream comparison is easily affected by page reflow, element relocation, and local layout reorganization. Removing type-specific representations also causes broad performance drops, with D-F1, S-Sim, and M-F1 decreasing to 0.811, 0.718, and 0.784, respectively, indicating that heterogeneous document elements cannot be effectively compared using a single generic representation space.

The two structure-related variants further clarify the role of structural modeling. When structural compatibility is removed from the alignment score, A-Acc and M-F1 decrease to 0.812 and 0.818, while S-Sim and S-Cons decline to 0.779 and 0.802. This shows that structural cues help not only difference reasoning but also the establishment of reliable cross-version correspondence. In contrast, removing structure-aware reasoning mainly affects structure-related metrics, reducing S-Sim from 0.842 to 0.703 and S-Cons from 0.872 to 0.728, while A-Acc and M-F1 remain unchanged because the alignment module is kept fixed. Using shared coefficients instead of type-specific coefficients also reduces all metrics, confirming that text, tables, formulas, and figures rely on location, content, and structure cues to different extents. Finally, removing typography cues mainly affects structure-related scores, with S-Sim and S-Cons decreasing to 0.815 and 0.841, demonstrating that font style, font size, and superscript/subscript information are meaningful revision signals in scientific text differencing.

\begin{table}[t]
\centering
\begin{threeparttable}
\caption{Core module-level ablation results of the proposed framework. Scores are macro-averaged over applicable element types.}
\label{tab:ablation_core}
\footnotesize
\setlength{\tabcolsep}{3.1pt}
\renewcommand{\arraystretch}{1.2}
\begin{tabular}{lccccccc}
\toprule
\textbf{Ablation} & \textbf{D-F1} & \textbf{\, L-F1 \,} & \textbf{IoU} & \textbf{S-Sim} & \textbf{S-Cons} & \textbf{A-Acc} & \textbf{M-F1} \\
\midrule
w/o type rep.        & 0.811 & 0.754 & 0.671 & 0.718 & 0.742 & 0.779 & 0.784 \\
w/o alignment        & 0.779 & 0.706 & 0.628 & 0.690 & 0.713 & 0.681 & 0.692 \\
w/o str. comp.       & 0.833 & 0.789 & 0.714 & 0.779 & 0.802 & 0.812 & 0.818 \\
w/o str. reasoning \;   & 0.829 & 0.776 & 0.709 & 0.703 & 0.728 & 0.856 & 0.859 \\
w/o type coeff.      & 0.842 & 0.803 & 0.726 & 0.813 & 0.831 & 0.827 & 0.832 \\
w/o typo. cues       & 0.857 & 0.818 & 0.749 & 0.815 & 0.841 & 0.850 & 0.854 \\
\midrule
\rowcolor[HTML]{ECF0FF}
\textbf{Full model}  & \textbf{0.866} & \textbf{0.833} & \textbf{0.749} & \textbf{0.842} & \textbf{0.872} & \textbf{0.856} & \textbf{0.859} \\
\bottomrule
\end{tabular}
\begin{tablenotes}[flushleft]
\footnotesize
\item D-F1, L-F1, S-Sim, S-Cons, A-Acc, and M-F1 denote detection F1, localization F1, structure similarity, structure consistency, alignment accuracy, and matching F1, respectively. Scores are macro-averaged over applicable element types. IoU is averaged over table and figure elements. The variant ``w/o str. reasoning'' keeps the same alignment module as the full model.
\end{tablenotes}
\end{threeparttable}
\end{table}

\noindent
\textbf{Alignment ablation.}
Fig.~\ref{fig:Figure4} analyzes the contribution of different compatibility terms in cross-version element alignment. Here, \textit{Fixed order} aligns elements according to page order and reading order without using the proposed compatibility score; \textit{w/o loc.}, \textit{w/o cont.}, and \textit{w/o str.} remove the location, content, and structural compatibility terms from the alignment score, respectively. As shown in Fig.~\ref{fig:Figure4}(a), fixed-order matching performs the worst, with an M-F1 of only 0.692, confirming that sequential or page-order correspondence is unreliable for production proofreading documents. In contrast, the full alignment model achieves the highest M-F1 of 0.859, showing that jointly modeling location, content, and structural compatibility provides more reliable matched pairs for downstream differencing.

The element-wise alignment results in Fig.~\ref{fig:Figure4}(b) further show that different compatibility terms contribute differently across element types. Removing the content term causes the largest degradation for text and figure elements, reducing their alignment accuracy from 0.914 to 0.804 and from 0.828 to 0.755, respectively. This indicates that semantic or visual similarity is essential for matching elements after layout changes. Removing the structural term mainly affects tables and formulas, where alignment accuracy decreases from 0.846 to 0.778 and from 0.835 to 0.755, respectively, confirming the importance of row--column topology, merged-cell relations, operator hierarchy, and superscript/subscript dependencies. Removing the location term has the most visible impact on figures, reducing alignment accuracy from 0.828 to 0.772, which shows that spatial proximity remains useful for visually grounded elements. Overall, the full model achieves the highest macro A-Acc of 0.856, confirming that location, content, and structure provide complementary alignment cues.

\subsection{Parameter sensitivity analysis}

\noindent
\textbf{Effect of compatibility weight profiles.}
We first analyze the influence of compatibility weight profiles in the alignment score defined in Eq.~\eqref{eq4}. In the full model, the alignment score uses type-specific coefficients $\lambda_{k,1}$, $\lambda_{k,2}$, and $\lambda_{k,3}$ for different semantic element types, reflecting the fact that text, tables, formulas, and figures rely on location, content, and structural compatibility to different extents. To examine whether such type-specific weighting is necessary, we construct several shared-weight profiles in which the same global weights are applied to all element types. These shared profiles include location-heavy $(0.60,0.20,0.20)$, content-heavy $(0.20,0.60,0.20)$, structure-heavy $(0.20,0.20,0.60)$, and uniform $(1/3,1/3,1/3)$ settings, where the three values correspond to the global weights of location, content, and structural compatibility, respectively.

\begin{figure}[t]
    \centering
    \includegraphics[width=1\linewidth]{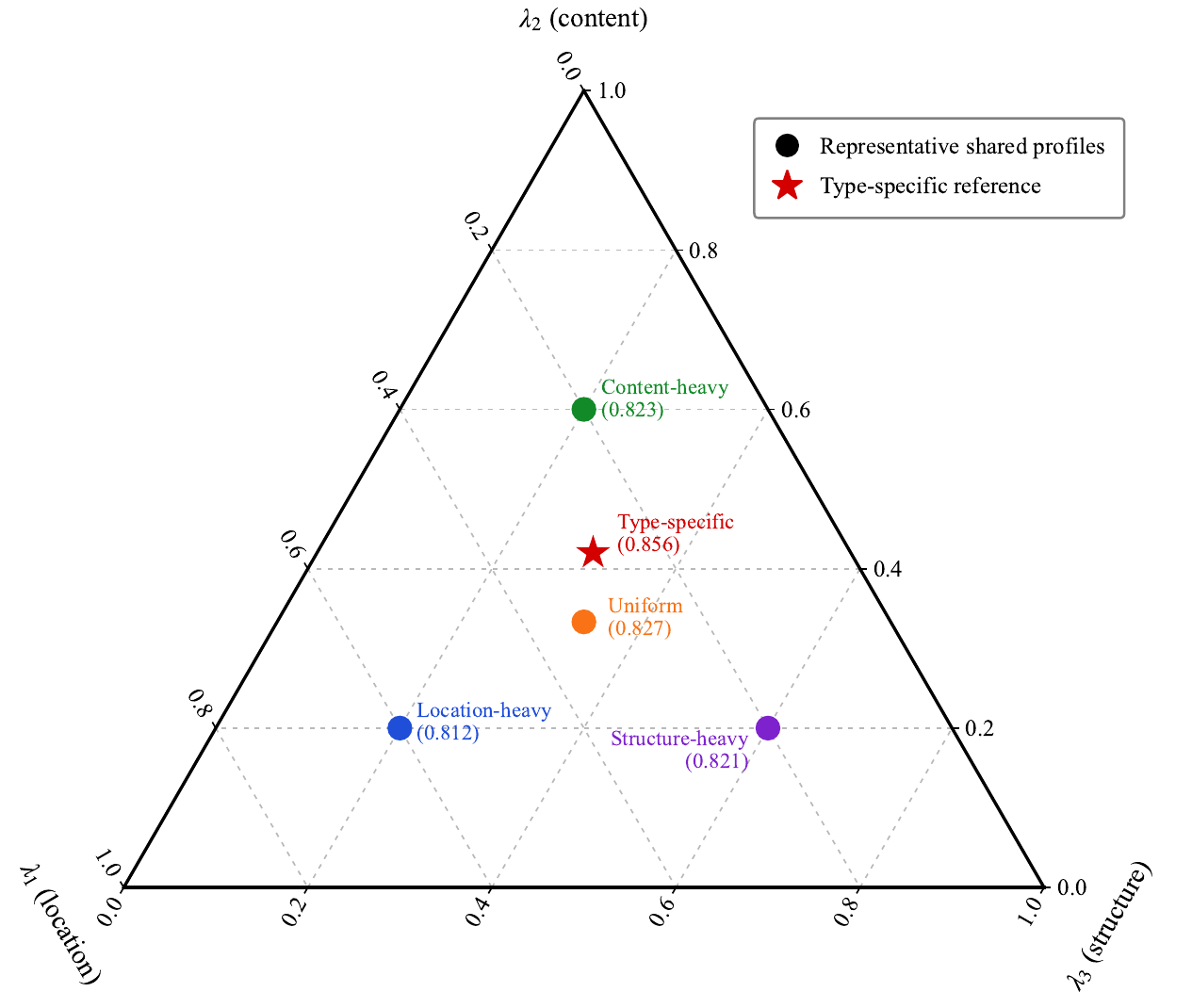}
\vspace{-.6cm}
\caption{Sensitivity analysis of compatibility weight profiles. The ternary scatter plot shows representative shared-weight profiles satisfying $\lambda_{1}+\lambda_{2}+\lambda_{3}=1$, where $\lambda_{1}$, $\lambda_{2}$, and $\lambda_{3}$ denote the global weights of location, content, and structural compatibility, respectively. The default type-specific setting, which uses element-dependent coefficients $\lambda_{k,1}$, $\lambda_{k,2}$, and $\lambda_{k,3}$, is shown only as a reference.}
\label{fig:Figure5}
\end{figure}

As shown in Fig.~\ref{fig:Figure5}, the default type-specific setting achieves the best performance, with an A-Acc of 0.856 and an M-F1 of 0.859, which are consistent with the full model results in the main comparison table. Among the shared-weight profiles, the uniform setting obtains the strongest result, with an A-Acc of 0.827 and an M-F1 of 0.832, indicating that a balanced global combination of location, content, and structural cues is more reliable than a single-cue-dominant global weighting scheme. The location-heavy profile performs the worst, with an A-Acc of 0.812 and an M-F1 of 0.818, because production proofreading documents often contain page reflow and element relocation, making spatial proximity alone insufficient. Content-heavy and structure-heavy profiles achieve A-Acc/M-F1 values of 0.823/0.829 and 0.821/0.827, respectively, but still underperform the type-specific setting. These results confirm that heterogeneous document elements require different compatibility compositions: text and figures benefit more from content-sensitive matching, whereas tables and formulas depend more strongly on structural compatibility.

\noindent
\textbf{Effect of alignment confidence thresholds.}
We next analyze the influence of the alignment confidence threshold $\tau_k$, which determines whether the best-scored candidate is accepted as a valid cross-version match. Since the default thresholds are type-specific, we apply the same relative perturbation $\Delta\tau$ to all element-specific thresholds while keeping the other settings unchanged. As shown in Table~\ref{tab:sensitivity_threshold}, the default setting $\Delta\tau=0$ achieves the best overall balance, with the highest Det-F1, Loc-F1, A-Acc, and M-F1 scores of 0.866, 0.833, 0.856, and 0.859, respectively. This indicates that the validation-selected thresholds provide a suitable acceptance boundary for distinguishing reliable matches from ambiguous candidates.

The results reveal a clear precision--recall trade-off in matching. When the thresholds are reduced, more candidate pairs are accepted, which helps preserve recall but also introduces more false correspondences. For example, at $\Delta\tau=-0.10$, M-R remains relatively high at 0.854, but M-P drops to 0.796, leading to a lower M-F1 of 0.824. Such false matches further affect downstream differencing, as changes may be assigned to incorrect element pairs. Conversely, increasing the thresholds makes alignment more conservative. At $\Delta\tau=+0.10$, M-P increases to 0.872, but M-R decreases to 0.803 because more true matches are rejected as unmatched, which also weakens Det-F1 and Loc-F1. The small performance variation at $\Delta\tau=\pm0.05$ indicates that the framework is stable around the default thresholds, which provide the best trade-off between false matches and missed correspondences.

\begin{figure}[t]
    \centering
    \includegraphics[width=1\linewidth]{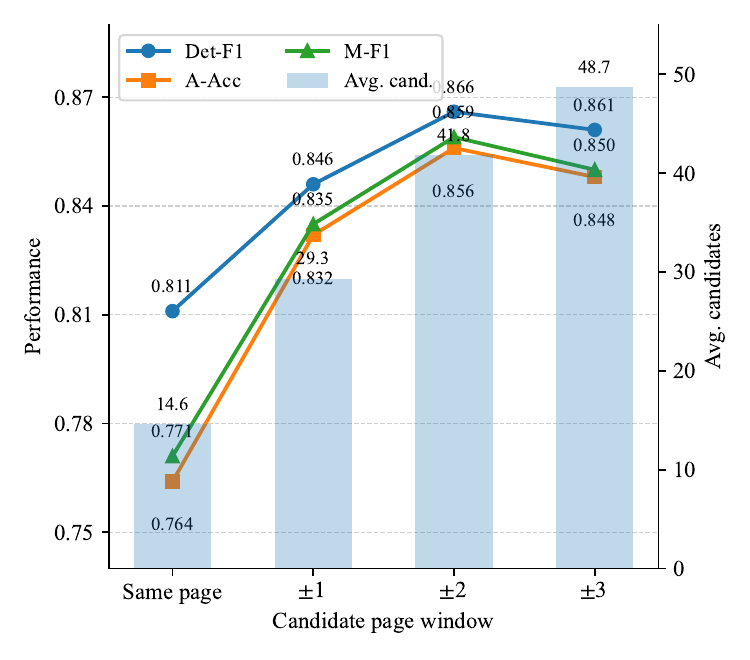}
\vspace{-.9cm}
\caption{Effect of candidate page window on cross-version alignment. The left y-axis shows Det-F1, A-Acc, and M-F1, while the right y-axis shows the average number of candidate elements considered for each source element.}
\label{fig:Figure6}
\end{figure}

\noindent
\textbf{Effect of candidate page window.}
We further analyze the influence of the candidate page window used in cross-version alignment. The page window controls the range within which candidate elements are retrieved before compatibility scoring, and is therefore closely related to page reflow, element relocation, and pagination shifts in production proofreading documents. Fig.~\ref{fig:Figure6} reports the results under four settings: same-page matching, $\pm1$ page, $\pm2$ pages, and $\pm3$ pages. The $\pm2$-page setting corresponds to the default configuration used in the main experiments and achieves the best overall performance, with a Det-F1 of 0.866, an A-Acc of 0.856, and an M-F1 of 0.859.

The results show a clear trade-off between candidate coverage and matching ambiguity. Same-page matching gives the weakest performance, with an A-Acc of 0.764 and an M-F1 of 0.771, because it cannot cover cross-page element displacement caused by pagination or layout reorganization. Expanding the window to $\pm1$ page improves A-Acc and M-F1 to 0.832 and 0.835, respectively, while the $\pm2$-page setting further improves them to 0.856 and 0.859. However, enlarging the window to $\pm3$ pages slightly reduces A-Acc and M-F1 to 0.848 and 0.850, respectively, because more irrelevant candidates are introduced. Meanwhile, the average number of candidates increases monotonically from 14.6 under same-page matching to 48.7 under the $\pm3$-page setting. These results indicate that the $\pm2$-page window provides the best balance between covering realistic cross-page revisions and controlling candidate noise.

\begin{table}[t]
\centering
\begin{threeparttable}
\caption{Sensitivity analysis of alignment confidence thresholds.}
\label{tab:sensitivity_threshold}
\footnotesize
\setlength{\tabcolsep}{7pt}
\renewcommand{\arraystretch}{1.15}
\begin{tabular}{lcccccc}
\toprule
$\boldsymbol{\Delta\tau}$ 
& \textbf{Det-F1} 
& \textbf{Loc-F1} 
& \textbf{A-Acc} 
& \textbf{M-P} 
& \textbf{M-R} 
& \textbf{M-F1} \\
\midrule
$-0.10$ & 0.842 & 0.806 & 0.821 & 0.796 & 0.854 & 0.824 \\
$-0.05$ & 0.858 & 0.824 & 0.846 & 0.837 & 0.861 & 0.849 \\
$0$     & \textbf{0.866} & \textbf{0.833} & \textbf{0.856} & 0.858 & \textbf{0.860} & \textbf{0.859} \\
$+0.05$ & 0.861 & 0.827 & 0.851 & 0.866 & 0.839 & 0.852 \\
$+0.10$ & 0.847 & 0.811 & 0.833 & \textbf{0.872} & 0.803 & 0.836 \\
\bottomrule
\end{tabular}
\begin{tablenotes}[flushleft]
\footnotesize
\item $\Delta\tau$ denotes the relative perturbation applied to the default type-specific alignment thresholds $\tau_k$. The default thresholds for text, table, formula, and figure elements are 0.72, 0.66, 0.70, and 0.64, respectively. A-Acc, M-P, M-R, and M-F1 denote alignment accuracy, matching precision, matching recall, and matching F1, respectively.
\end{tablenotes}
\end{threeparttable}
\end{table}

\section{Discussion and Implications}
\label{discussion}

The experimental results suggest that cross-version scientific document differencing should be understood as a heterogeneous and alignment-dependent reasoning problem rather than as a conventional text-diff or image-diff task. Text-oriented methods can capture local content edits, but they are sensitive to page reflow, paragraph movement, and typography-sensitive revisions. Visual comparison methods can highlight apparent page-level differences, but they provide limited semantic interpretability and are vulnerable to rendering variation. In contrast, the proposed framework performs differencing over aligned semantic elements, which explains its consistent improvements across change detection, localization, structure awareness, and alignment/matching quality.

A key observation is that alignment is central to reliable scientific document comparison. The ablation results show that removing cross-version alignment causes the largest degradation, with alignment accuracy decreasing from 0.856 to 0.681 and matching F1 decreasing from 0.859 to 0.692. This confirms that many apparent differences are caused by displacement, pagination changes, or local layout reorganization rather than true content revisions. The alignment ablation further shows that location, content, and structural compatibility contribute differently across element types: content similarity is more important for text and figures, structural compatibility is more critical for tables and formulas, and location information remains useful for visually grounded elements. These findings support type-specific alignment, as different elements rely on distinct spatial, content, and structural cues.

The results also highlight the importance of structure-aware reasoning. Tables and formulas cannot be adequately compared using flattened text or visual appearance alone, because their meanings depend on row--column topology, merged-cell relations, operator hierarchy, and superscript/subscript dependencies. The proposed method improves table structure similarity from 0.716 to 0.833 over the strongest table baseline and formula structure consistency from 0.797 to 0.877 over the strongest formula baseline. Similarly, the improvement on typography-sensitive text revisions indicates that font size, bold/italic style, and superscript/subscript status can carry structural or semantic implications in scientific documents.

From a practical perspective, the framework is well aligned with real editorial production workflows. Scientific publishing involves multiple proofreading rounds, where revisions may include content edits, layout adjustment, table reformatting, formula modification, figure replacement, and typography-sensitive changes. The scenario-level analysis shows that the proposed method remains effective under representative production proofreading conditions, including text reflow, table reformatting, formula hierarchy revision, figure replacement, and local figure updates. These results indicate that the framework can support interpretable change verification and reduce manual inspection effort in scholarly publishing workflows.

Several limitations remain. First, the current framework primarily assumes one-to-one element correspondence, which may be insufficient for paragraph splitting/merging, table decomposition across pages, or substantial layout restructuring. Second, structure-aware differencing depends on upstream parsing quality; errors in table cell reconstruction, formula recognition, or figure-region extraction may propagate to downstream reasoning. Third, figure differencing is currently evaluated mainly through region-level localization because large-scale component-level structural annotations are unavailable in the production dataset. Finally, although the framework is designed for scientific documents in general, the experiments are conducted mainly on scientific PDF versions from real proofreading workflows. Future work will explore many-to-many and graph-based alignment, stronger structured element parsers, richer figure component annotations, and extension to broader publishing formats such as Word, XML, HTML, and LaTeX-based workflows.

\section{Conclusion}
\label{conclusion}

This paper addressed cross-version differencing of scientific documents, highlighting the limitations of conventional text-sequence-based and image-based comparison methods. We formulated the problem as a unified task of heterogeneous element decomposition, cross-version alignment, and structure-aware difference reasoning. Based on this formulation, we proposed a layout-aware and heterogeneous element-aware framework that decomposes document versions into text, tables, formulas, and figures, establishes stable cross-version correspondences, and performs type-aware differencing over aligned element pairs. Experiments on real-world production proofreading data show consistent improvements over element-specific baselines across all evaluation dimensions.

The results also confirm that type-specific representations, alignment-first design, structure-aware reasoning, and compatibility-weight modeling are critical for reliable and interpretable differencing. Parameter sensitivity and ablation analyses show that the framework is robust to variations in weights, thresholds, and candidate windows. Overall, the proposed approach provides a practical and interpretable solution for analyzing revisions in complex scientific documents and lays the foundation for future extensions, including many-to-many correspondence modeling, improved structured element parsing, and cross-format document differencing across diverse publishing workflows.

\bibliographystyle{elsarticle-num} 
\bibliography{ref}

\end{document}